\documentclass[conference]{IEEEtran}
\IEEEoverridecommandlockouts
\usepackage{cite}
\usepackage{amsmath,amssymb,amsfonts}
\usepackage{algorithm}
\usepackage{algorithmicx}
\usepackage{algpseudocode} 
\usepackage{graphicx}
\usepackage{textcomp}
\usepackage{xcolor}
\usepackage{url}
\usepackage{subfigure}
\usepackage{booktabs}
\usepackage{multicol}
\usepackage{multirow}
\def\BibTeX{{\rm B\kern-.05em{\sc i\kern-.025em b}\kern-.08em
    T\kern-.1667em\lower.7ex\hbox{E}\kern-.125emX}}

\begin{document}

\title{Generating Adversarial Examples with Better Transferability via Masking Unimportant Parameters of Surrogate Model}

\author{\IEEEauthorblockN{Dingcheng Yang$^{1,2}$, Wenjian Yu$^{1,*}$\thanks{$^*$corresponding author.}, Zihao Xiao$^2$, Jiaqi Luo$^1$}
\IEEEauthorblockA{$^1$Dept. Computer Science \& Tech., BNRist, Tsinghua University, Beijing, China.  \\
$^2$RealAI. \\
ydc19@mails.tsinghua.edu.cn, yu-wj@tsinghua.edu.cn, zihao.xiao@realai.ai, luojq19@mails.tsinghua.edu.cn
}
}

\maketitle

\begin{abstract}
Deep neural networks (DNNs) have been shown to be vulnerable to adversarial examples. Moreover, the transferability of the adversarial examples has received broad attention in recent years, which means that adversarial examples crafted by a surrogate model can also attack unknown models. This phenomenon gave birth to the transfer-based adversarial attacks, which aim to improve the transferability of the generated adversarial examples. In this paper, we propose to improve the transferability of adversarial examples in the transfer-based attack via masking unimportant parameters (MUP). The key idea in MUP is to refine the pretrained surrogate models to boost the transfer-based attack. Based on this idea, a Taylor expansion-based metric is used to evaluate the parameter importance score and the unimportant parameters are masked during the generation of adversarial examples. This process is simple, yet can be naturally combined with various existing gradient-based optimizers for generating adversarial examples, thus further improving the transferability of the generated adversarial examples. Extensive experiments are conducted to validate the effectiveness of the proposed MUP-based methods. 
\end{abstract}

\section{Introduction}
\begin{figure}[h]
  \setlength{\abovecaptionskip}{0.1 cm}
  \centering
  \subfigure[Original Image.]
    {
    \includegraphics[width=1.6in]{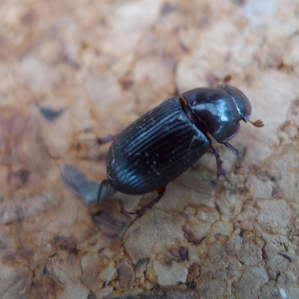}
    }
    \subfigure[Advdersarial Example.]
    {
    \includegraphics[width=1.6in]{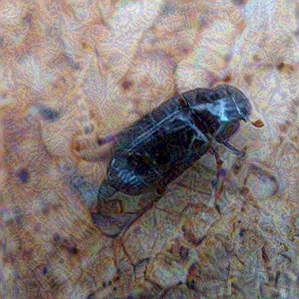}
    }
    \caption{A visualization of adversarial examples. The left image is a clean image of a dung beetle, and the right image is an adversarial example generated from an Inception-v3 model using the method proposed in this paper, and which is misclassified as a dragonfly.}
      \label{fig:adv}
\end{figure}

Deep neural networks (DNNs) have achieved remarkable success in many areas, such as computer vision and natural language processing. However, they are vulnerable to adversarial examples, which are generated by adding carefully designed imperceptible perturbations on clean data. An example is illustrated in Fig.~\ref{fig:adv}. The left image is an image of a dung beetle, which is also successfully classified as a dung beetle by an Inception-v3 network~\cite{szegedy2016rethinking}. However, a carefully constructed adversarial example (the right image) fools Inception-v3 to misclassify it as a dragonfly, even though it does not look any different from the left image to a human. The generation of adversarial samples is known as the adversarial attack, which raises serious concerns about the security of DNN for deployments in real-world scenarios, such as face recognition and autonomous driving.

Even worse, researchers have found that adversarial examples have transferability. For example, although the adversarial example in Fig.~\ref{fig:adv} was generated from an Inception-v3, it can also successfully fool an Inception-v4 model. Such a phenomenon allows attacking unknown victim models by attacking a given surrogate model, which can be obtained by training locally or by using a publicly available pretrained model. Generating adversarial examples from a given model is called the \textit{white-box attack}. In contrast, the attack that generates adversarial examples from a surrogate model and leverages its transferability to fool unknown victim models is called the \textit{transfer-based attack}. The transfer-based attack is more practical than the white-box attack due to the inaccessibility of real-world commercial models.

To better verify the safety of a DNN, investigating the method of transfer-based attack has received widespread attention. Dong et al. were the first to treat the problem of generating adversarial examples as a model training problem~\cite{dong2018boosting}. Specifically, the surrogate models were treated as the training data and the victim models were treated as the testing data, so the transferability of adversarial examples was naturally regarded as the generalizability of DNNs. Inspired by this analogy, many ideas to alleviate the overfitting phenomenon were borrowed to propose new methods for the transfer-based attack. Dong et al.~\cite{dong2018boosting} proposed a momentum iterative gradient-based method to escape from pool local maxima and experimentally improved the transferability. Xie et al.~\cite{xie2019improving} proposed to use data augmentations to escape local optima. Then, there are a series of works that improve the transferability by using different data augmentation methods, including translation~\cite{dong2019evading}, Mixup~\cite{wang2021admix}, and scaling~\cite{lin2019nesterov}. More recently, Huang et al.~\cite{huang2022transferable} proposed an algorithm named Transferable Attack based on Integrated Gradients on Random piece-wise linear path (TAIG-R) to avoid overfitting the surface of the surrogate model.

This work is also inspired by~\cite{dong2018boosting}, but instead of proposing an optimizer to avoid overfitting a given surrogate model, we want to refine the given surrogate model thus boosting the transfer-based attack. This is similar to the data cleaning process of model training, where the model performance is improved by removing noisy data. Our motivation comes from the phenomenon observed in the field of network pruning, where a DNN often has a large number of unimportant parameters. We propose to mask these unimportant parameters when generating adversarial examples to avoid overfitting them. We first borrow the idea from the work on model pruning~\cite{molchanov2017pruning}, using a Taylor expansion-based metric to evaluate the parameter importance scores of the surrogate model. Then, we propose a technique for boosting transfer-based attacks which improves the transferability of adversarial examples via masking unimportant parameters (MUP), where the unimportant parameters are identified by the Taylor expansion-based metric. The experimental results on ImageNet-compatible dataset show that the proposed method can consistently improve the transferability of adversarial examples.

Note that we are not the only work to mask model parameters during the generation of adversarial examples. The ghost network (GN)~\cite{li2020learning} leveraged dropout layers~\cite{srivastava2014dropout} to prevent overfitting,  which is equivalent to randomly masking some neurons. The key difference in our method is the way to identify the parameters to be masked, i.e., we use a Taylor expansion-based metric to identify the unimportant parameters rather than randomly dropout. Experimental comparison with the method~\cite{li2020learning} will be presented in Section IV.B.

The major contributions of this work are as follows:

\begin{itemize}
    \item We propose the idea of improving the transferability of adversarial examples via masking unimportant parameters (MUP) in the surrogate model.
    \item A technique for boosting transfer-based attacks is presented based on the proposed idea, which utilizes a Taylor expansion-based metric to evaluate the parameter importance score of surrogate models, and then masks unimportant parameters to improve existing optimizers for generating adversarial examples.
    \item Extensive experiments on ImageNet-compatible dataset demonstrate the effectiveness of the proposed method. With the TAIG-R optimizer~\cite{huang2022transferable}, the proposed method improves the attack success rates by 4.9\%, 4.2\%, and 2.9\% on average respectively when Inc-v3, Inc-v4, and DN121 are used as surrogate models. Moreover, the experimental results also show that the proposed method is better than the comparative work GN~\cite{li2020learning}.
\end{itemize}

\section{Related Work}
\subsection{Adversarial Attack}
Although DNNs have demonstrated to be successful on many tasks, they are vulnerable to adversarial examples~\cite{szegedy2013intriguing}. Let $x \in \mathbb{R}^d$ denote an image with dimension $d$, $y$ denote its ground truth label. Given a pretrained DNN $\theta$, the adversarial attack is implemented by solving the following optimization problem: 
\begin{equation}
    \begin{split}
        & x^* = \text{argmax}_{x'} \mathcal{L}(x',y,\theta)~, \\
        & \text{s.t.}\quad \|x'-x\|_{\infty} \le \epsilon~,
        \label{eq:eq1}
    \end{split}
\end{equation}
where the $\mathcal{L}(x,y,\theta)$ denotes the loss function that guides the DNN $\theta$ to predict the class of image $x$ as ground-truth $y$, $\epsilon$ is the maximum allowed magnitude of perturbation, and $x^*$ is the generated adversarial example. Some gradient-based optimizers were proposed to solve~\eqref {eq:eq1}. Szegedy et al. were the first to use a box-constrained L-BFGS optimizer~\cite{szegedy2013intriguing}. Then, a one-step optimizer, i.e., fast gradient sign method (FGSM) was proposed to efficiently generate adversarial examples~\cite{goodfellow2014explaining}. It was later extended to the iterative version of FGSM (I-FGSM)~\cite{kurakin2018adversarial}, which generates adversarial examples that fool the model $\theta$ with a higher success rate.

The generated adversarial example is shown to have the ability to attack an unknown model, and this ability is called the transferability of the adversarial example. In this scenario, the model $\theta$ used to generate the adversarial example is called the \textit{surrogate model}, and the unknown model to be attacked is called the \textit{victim model}. This type of adversarial attack is called the transfer-based attack. Dong et al. systematically investigated the transferability of adversarial examples and found an interesting phenomenon that is although the iterative method (I-FGSM) is a stronger optimizer than the one-step method (FGSM), the transferability of the adversarial examples generated by I-FGSM is worse than FGSM~\cite{dong2018boosting}. Specifically, their experimental results on the white-box attacks show that an Inception-v3 has only a 72.3\% probability of being fooled by FGSM, but a 100\% probability of being fooled by I-FGSM. However, when the adversarial examples generated from Inception-v3 were then used to attack an Inception-v4 model, the attack success rates of the adversarial examples generated by FGSM and I-FGSM were 28.2\% and 22.8\%, respectively. It shows that the adversarial examples generated by FGSM are more transferable. 

To explain this phenomenon, Dong et al. analogized the surrogate model to the training data and the victim model to the testing data, so the transferability of the generated adversarial examples can be analogous to the generalizability of the trained models. Then, inspired by the widespread use of momentum-based optimizers to escape local maxima in training DNNs, Dong et al. propose a Momentum Iterative fast gradient sign Method (MIM) to generate more transferable adversarial examples~\cite{dong2018boosting}. Suppose the number of iterations is $N$, the MIM performs the following steps in an iteration:
\begin{align}
    &\delta_{t+1} = \nabla_x  \mathcal{L}(x_t,y,\theta)~,\\
    &g_{t+1} = \mu \cdot g_t + \delta_{t+1}/\|\delta_{t+1}\|_1 ~,\\
    &x_{t+1} = \textrm{Clip}_x^{\epsilon} \{ x_t+\beta \cdot \textrm{sign}(g_{t+1}) \}~, ~ ~ ~ t=0, 1, \cdots
\end{align}
with $x_0=x$ and $g_0=\mathbf{0}$, and the generated adversarial example is $x_N$. Here, the $\textrm{Clip}_x^{\epsilon}$ function is used to guarantee that the adversarial example is in the $\epsilon$-ball of $x$ under the $L_{\infty}$ norm, $\mu$ denotes a momentum factor and $\beta$ is a step size. 

Furthermore, many optimizers have been proposed to improve MIM by introducing data augmentation to the objective function $\mathcal{L}(x',y,\theta)$ in~\eqref{eq:eq1} to avoid overfitting. For example, Lin et al. proposed a Scale-Invariant attack Method (SIM)~\cite{lin2019nesterov}, which optimizes an adversarial example over a set of scaled images. Specifically, the loss function for generating adversarial examples with SIM is $\sum_{i=0}^{m-1} \mathcal{L}(x/2^i,y,\theta)$, where $m$ denotes the number of scaled images. More recently, Huang et al. proposed an algorithm named Transferable Attack based on Integrated Gradient (TAIG)~\cite{huang2022transferable}, which computes the gradient over a set of sampling images. Two versions of TAIG were proposed, the stronger version is called TAIG-R, which is equivalent to optimizing the following loss function: $\sum_{i=1}^S \mathcal{L}(\frac{i}{S}x+U(-\epsilon,\epsilon),y,\theta)$, where $U$ denotes a uniform distribution, and $S$ denotes the number of sampling images.

Unlike the above methods, instead of proposing a new optimizer to generate adversarial examples (by solving~\eqref{eq:eq1}), we propose the idea of masking unimportant parameters in the surrogate model $\theta$ when using them. Therefore, our approach can assist the above methods, as will be shown in Section III.B. To the best of our knowledge, the most relevant work to us is the ghost network (GN)~\cite{li2020learning}, which manually injects some dropout layers into the surrogate model to randomly drop some neurons. Our experimental results in Section IV.B will show that our approach works better than GN.

\subsection{Network Pruning}
Network pruning is a popular technique in the field of model compression, which aims to remove unimportant parameters to compress and accelerate DNNs. To guarantee the accuracy of the pruned model, many heuristic strategies have been proposed to evaluate whether a parameter is important or not. Optimal Brain Damage~\cite{lecun1989optimal} pruned the parameters based on the Hessian matrix of the loss function. Deep Compression~\cite{han2016deep} pruned the parameters based on their absolute values. Li et al.~\cite{li2017pruning} proposed an $L_1$-norm-based metric to evaluate the importance of channels. Molchanov et al.~\cite{molchanov2017pruning} proposed a Taylor expansion-based metric to evaluate the importance of channels. In this paper, we borrow the idea from the work~\cite{molchanov2017pruning} to evaluate the parameter importance scores based on the first-order gradient information of the parameters.
\section{Method}
In this section, we first present a Taylor expansion-based metric for calculating the parameter importance score. Based on this metric, we present a technique for boosting transfer-based attacks that masks the unimportant parameters at each iteration.
\subsection{Parameter Importance Score}
Existing transfer-based attack methods generate adversarial examples by maximizing the loss function~\eqref{eq:eq1} and improving the transferability of the adversarial examples by using advanced optimizers that help to escape the local maxima. The parameter $\theta$ in~\eqref{eq:eq1} is considered as a constant in these works. Inspired by the analogy of surrogate models to training data in~\cite{dong2018boosting}, we propose that the quality of $\theta$ is also a key factor for the transferability, just as noisy data can be detrimental to generalizability.

However, work on network pruning suggests that DNNs are often over-parameterized. For example, Li et al.~\cite{li2017pruning} showed that even after removing 50$\%$ of the convolutional kernels from the first convolutional layer of a VGG16 model~\cite{simonyan2014very} based on the $L_1$-norm, the accuracy degradation of the VGG16 model on CIFAR-10 dataset~\cite{krizhevsky2009learning} was still negligible. Since there are redundant parameters for prediction, it is natural to be concerned about the impact of these redundant parameters on the generation of adversarial examples. While previous work on preventing overfitting can alleviate this effect, explicitly removing these parameters is also a worthwhile consideration, and the two types of approaches can be naturally combined to achieve better performance. To this end, we first introduce a method to calculate the importance scores of model parameters in this subsection. Based on the calculated scores, we can explicitly identify and mask unimportant parameters during the generation of adversarial examples, which will be presented in the next subsection.

Given an image $x$ and its true label $y$, assume that $V_i$ represents the importance score of the $i$-th parameter $\theta_i$ for predicting the category of image $x$ as label $y$. It can be defined intuitively as $V_i=|\mathcal{L}(x,y,\theta)-\mathcal{L}(x,y,\theta-\theta_ie_i)|$, where $e_i$ denotes a one-hot vector consisting of zeros in all elements except for a single one in $i$-th element. It means how much the loss function change after setting $\theta_i$ to zero. However, calculating the importance scores for all parameters in this way requires a number of forward propagations proportional to the model size, which is intractable for large DNNs. To reduce the computational overhead, It was approximated in~\cite{molchanov2017pruning} by using the first-order gradient information based on the Taylor expansion as follows:
\begin{equation}
    \label{eq:V}
    V_i=|\mathcal{L}(x,y,\theta)-\mathcal{L}(x,y,\theta-\theta_ie_i)| \approx |\frac{\partial \mathcal{L}(x,y,\theta)}{\partial \theta_i} \theta_i| .
\end{equation}

According to~\eqref{eq:V}, the importance scores of all parameters can be calculated simultaneously in just one step of backpropagation.
\subsection{The MUP-Based Attack Algorithm}
\begin{figure*}[h]
  \setlength{\abovecaptionskip}{0.1 cm}
  \centering
    \includegraphics[width=7in]{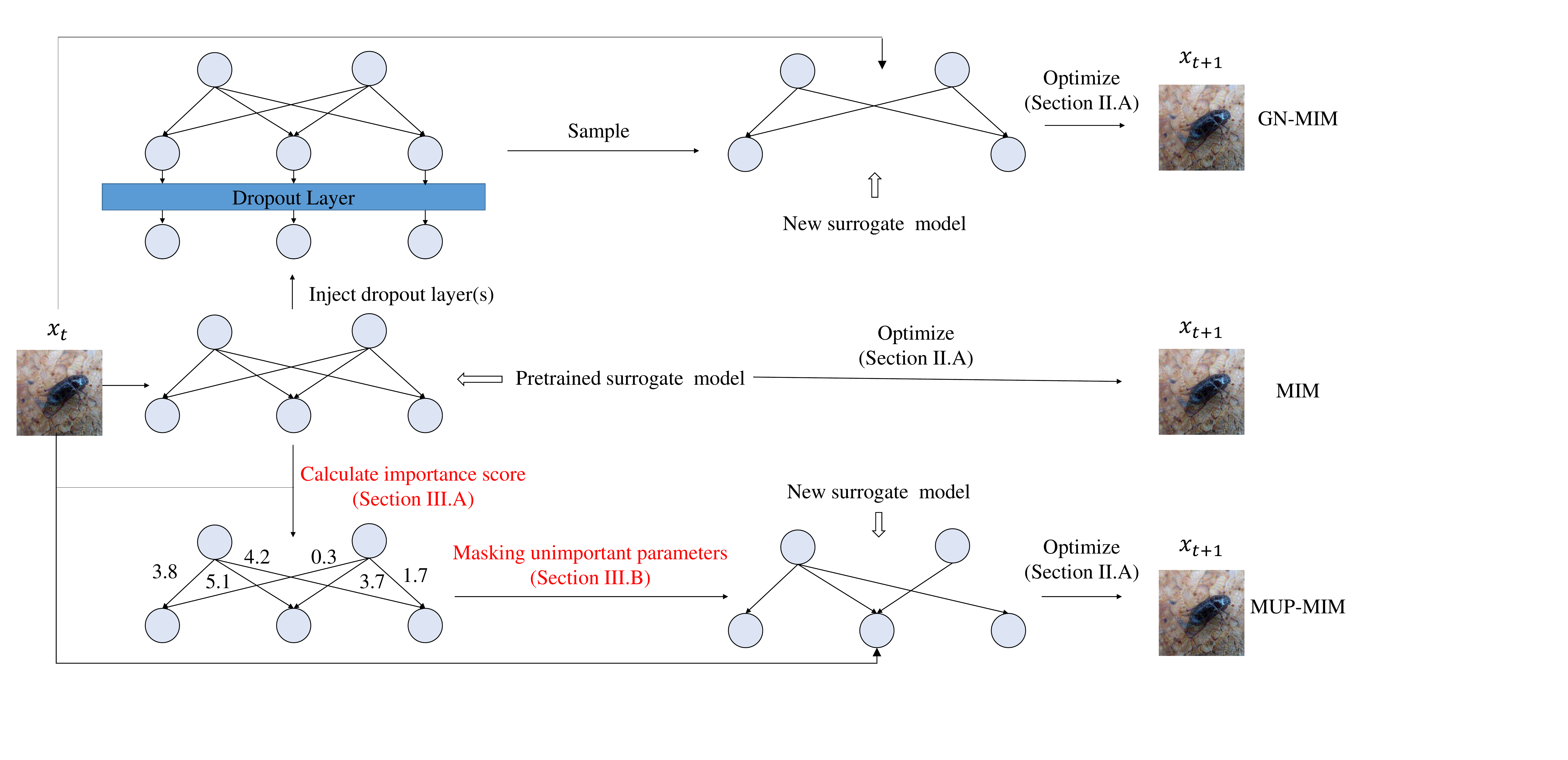}
  \caption{An illustration of the MUP technique and the approach of ghost network (GN) during a round of iteration for generating adversarial examples (taking the optimizer MIM as an example).
  }
    \label{fig:pipeline}
\end{figure*}

In this subsection, we propose a technique for boosting existing optimizers by masking unimportant parameters (MUP). We take assisting MIM as an example, call it MUP-MIM, and illustrate it in Fig.~\ref{fig:pipeline}. Only two layers of the surrogate models are drawn in Fig.~\ref{fig:pipeline} for simplicity. The standard MIM and the MIM assisted by the comparative work ghost network (GN)~\cite{li2020learning} are also illustrated in Fig.~\ref{fig:pipeline}, and the latter is called GN-MIM. As can be seen in Fig.~\ref{fig:pipeline}, the pretrained surrogate model is fixed in MIM, while it is changing dynamically when using GN and MUP.

Given a hyperparameter masking ratio $r$, MUP will mask unimportant parameters at each iteration. To do this, the important scores $V$ are calculated as described in Section III.A. Then, the $r \times |V|$ elements in $V$ with the smallest score are considered as unimportant parameters. This process is simple and can be naturally integrated into any gradient-based optimizer. We take the basic MIM as an example and summarize the algorithm MUP-MIM for assisting MIM with MUP in Algorithm~\ref{alg:MUP}. At each iteration, the importance scores are first calculated (Step 3). Then, a binary mask of the same shape as $\theta$ is obtained based on the given masking ratio $r$ (Step 4-5). The modification to the standard MIM is that the gradient vector $\delta_{t+1}$ is computed with the model $\theta \odot M$, whose unimportant parameters are masked (Step 6). Here the $\odot$ denotes an element-wise product. The subsequent steps are the same as for MIM (Step 7-8). Finally, the produced $x_N$ is the generated adversarial example. Since the unimportant parameters are masked at each iteration, the generated adversarial example avoids overfitting them and has higher transferability, as will be demonstrated in Section IV.B. 

\begin{algorithm}[h]
\caption{MUP-MIM:Masking unimportant parameters for assisting MIM.}\label{alg:MUP}
\begin{flushleft}
\textbf{Input:} Input image $x$ and its ground-truth label $y$, pretrained surrogate model $\theta$, number of iteration $N$, momentum factor $\mu$, step size $\beta$, the maximum allowed magnitude of perturbation $\epsilon$, masking ratio $r$.\\
\textbf{Output:} The generated adversarial example $x_N$ with $\|x_N - x\|_{\infty} \le \epsilon$
\end{flushleft}
\begin{algorithmic}[1]
\State $x_0 \gets x, g_0 \gets \mathbf{0}$.
\For{$t \gets 0$ to $N-1$}
\State $V \gets |\frac{\partial \mathcal{L}(x_t,y,\theta)}{\partial \theta} \odot \theta|$. \Comment{(5)}
\State Let $\tau$ be the $r \times |V|$-th smallest element in $V$.
\State $M \gets V > \tau$. \Comment{A binary mask.}
\State $\delta_{t+1} \gets \nabla_x \mathcal{L}(x_t,y,\theta \odot M)$. \Comment{(2)}
\State $g_{t+1} = \mu \cdot g_t + \delta_{t+1}/\|\delta_{t+1}\|_1$. \Comment{(3)}
\State $x_{t+1} = \textrm{Clip}_x^{\epsilon} \{ x_t+\beta \cdot \textrm{sign}(g_{t+1}) \}$. \Comment{(4)}
\EndFor
\end{algorithmic}
\end{algorithm}

It is straightforward to integrate MUP into other optimizers. Two improved versions of MIM are considered in this paper, namely SIM~\cite{lin2019nesterov} and TAIG-R~\cite{huang2022transferable}, which are introduced in Section II.A. Assisting SIM with MUP (called MUP-SIM), only the gradient $\delta_{t+1}$ in Step 6 needs to be computed as follows:
\begin{equation}
    \delta_{t+1} = \nabla_x \sum_{i=0}^{m-1} \mathcal{L}(x_t/2^i,y,\theta \odot M).
\end{equation}
Similarly, using MUP to assist TAIG-R (called MUP-TAIG-R) only requires to compute $\delta_{t+1}$ in Step 6 as follows:
\begin{equation}
    \delta_{t+1} = \nabla_x \sum_{i=1}^S \mathcal{L}(\frac{i}{S}x_t+U(-\epsilon,\epsilon),y,\theta \odot M).
\end{equation}

The comparative work ghost network (GN)~\cite{li2020learning} is also shown in Fig.~\ref{fig:pipeline}, which manually injects some dropout layers into the pretrained surrogate model to assist MIM. The authors of GN claim to do this in order to sample a sub-network at each iteration. And in our opinion, this is equivalent to randomly masking some \textit{neurons}. In contrast, MUP is selectively masking some parameters, which are called \textit{synapses} in computational neuroscience. Therefore, MUP has two advantages over GN in terms of the way to identify the parameters to be masked. Firstly, the proposed MUP is input-dependent and is able to model the effect of input data on the importance scores, and is therefore more theoretical. Secondly, MUP has a finer granularity in identifying unimportant parameters, since masking a neuron in GN is equivalent to masking a group of synapses. Besides the difference in the way to identify the parameters to be masked, GN needs to manually specify some locations to inject dropout layers, while not needed in MUP, so MUP is more convenient to be applied on a new architecture without any customization.

\section{Experimental Results}
In this section, we first provide our experimental setup in Section IV.A. Then, we demonstrate the effectiveness of our MUP on ImageNet-compatible dataset\footnote{https://www.kaggle.com/google-brain/nips-2017-adversarial-learning-development-set} in Section IV.B. Finally, we perform some ablation studies in Section IV.C to investigate the impact of different hyperparameters.
\subsection{Setup}
\textbf{Victim Models.}~We consider nine publicly available models as victim models, which have been widely used in previous work~\cite{dong2018boosting,lin2019nesterov}. The first three of them are normally trained models, including Inception-v3 (Inc-v3)~\cite{szegedy2016rethinking}, Inception-v4 (Inc-v4), and Inception-ResNet-v2 (IncRes-v2)~\cite{szegedy2017inception}. The rest are robust models: Inc-v3$_{ens3}$, Inc-v3$_{ens4}$, and IncRes-v2$_{ens}$~\cite{tramer2018ensemble}, high-level representation guided denoiser (HGD)~\cite{liao2018defense}, input transformation through resizing and padding (R\&P)~\cite{xie2017mitigating}, and the rank-3 submission in NIPS2017 adversarial competition (NIPS-r3)\footnote{https://github.com/anlthms/nips-2017/tree/master/mmd}.

\textbf{Surrogate Models.}~Following the setting of the comparative work ghost network (GN), we first choose Inc-v3 and Inc-v4 as the architectures of the surrogate model, since they were experimented in~\cite{li2020learning}. To demonstrate the versatility of MUP over more architectures, we additionally conduct experiments on the DenseNet121 (DN121) surrogate model.

\textbf{Optimizers.}~We use MUP to assist three optimizers for generating adversarial examples to evaluate the compatibility of the proposed MUP with prior works, including MIM~\cite{dong2018boosting}, SIM~\cite{lin2019nesterov}, and the state-of-the-art TAIG-R~\cite{huang2022transferable}. 

\textbf{Hyperparameters.}~ If not specified otherwise, the masking ratio $r$ is set to 
15\%, 30\% and 25\% for the surrogate model Inc-v3, Inc-v4, and DN121, respectively. Notice that it is independent with respect to the victim model since the victim model is unknown in the assumptions of the transfer-based attack. For the hyperparameters related to different optimizers, we follow the recommended values of their corresponding papers to set them. Specifically, we set the step size $\beta$ to 2, the number of iterations to 10, the momentum factor $\mu$ for MIM to 1.0, the number of scaled images $m$ for SIM to 5, and the number of sampling images $S$ for TAIG-R to 20. All the experiments are conducted with the maximum allowed magnitude of perturbation $\epsilon=16$.
\subsection{Results on ImageNet-compatible dataset}

In this subsection, we will first show that using the proposed MUP can improve the success rate of existing optimizers. In addition, we compared the proposed MUP with the comparative work ghost network (GN), which was used to assist MIM in~\cite{li2020learning}. To follow the experimental setup of GN, we first consider assisting MIM, as well as Inc-v3 and Inc-v4 as the architecture of the surrogate models. Then, a DN121 is additionally considered as a surrogate model, which reflects the versatility of MUP for different architectures of the surrogate model. In contrast, GN cannot be applied directly to a new architecture, as it requires specifically designed locations for adding dropout layers, as we present in Section III.B. All of the experiments are conducted on the ImageNet-compatible dataset used in the NIPS 2017 adversarial competition.

\begin{table*}[h]
  \setlength{\abovecaptionskip}{0.1 cm}
    \centering
    \caption{The attack success rates (\%) of different methods based on MIM~\cite{dong2018boosting}. ``-'' indicates the white-box attack.}
    \label{tab:MIM}
    \begin{tabular}{@{~}c@{~~}c@{~~}c@{~~}c@{~~}c@{~~}c@{~~}c@{~~}c@{~~}c@{~~}c@{~~}c@{~~}c@{~}}
        \toprule
        Surrogate Model & Method & Average & Inc-v3 & Inc-v4 & IncRes-v2 & Inc-v3$_{ens3}$ & Inc-v3$_{ens4}$ & IncRes-v2$_{ens}$& HGD & R\&D & NIPS-r3\\
         \cmidrule(r){1-1}
         \cmidrule(r){2-2}
         \cmidrule(r){3-12}
         \multirow{3}*{Inc-v3} & MIM & 23.4 & - & 54.1 & 48.1 & 21.8 & 21.3 & 10.7 & 6.5 & 10.0 & 14.4 \\
          & GN-MIM & 26.2 & - & 61.6 & 58.9 & 23.1 & 22.4 & 11.6 & 6.6 & 10.3 & 15.2 \\
          & MUP-MIM & \textbf{28.5} & - & \textbf{66.2} & \textbf{64.4} & \textbf{25.1} & \textbf{23.2} & \textbf{12.8} & \textbf{6.9} & \textbf{12.6} & \textbf{16.4}  \\
          \midrule
         \multirow{3}*{Inc-v4} & MIM & 23.8 & 64.2 & - & 49.9 & 18.3 & 18.6 & 10.3 & 6.0 & 9.9 & 12.9  \\
          & GN-MIM & 29.0 & 75.2 & - & 63.3 & 23.7 & 22.0 & \textbf{12.8} & 5.4 & \textbf{12.3} & \textbf{16.9}  \\
          & MUP-MIM & \textbf{29.6} & \textbf{76.8} & - & \textbf{65.4} & \textbf{24.0} & \textbf{23.0} & 12.4 & \textbf{7.3} & 11.8 & 16.4   \\
          \midrule
         \multirow{2}*{DN121} & MIM & 48.6 & 67.6 & 61.4 & 55.7 & 47.9 & 46.4 & 36.4 & 46.9 & 35.7 & 39.8 \\
          & MUP-MIM & \textbf{59.9} & \textbf{82.3} & \textbf{76.0} & \textbf{69.5} & \textbf{58.3} & \textbf{56.4} & \textbf{44.1} & \textbf{57.8} & \textbf{44.9} & \textbf{49.6}   \\
        \bottomrule
    \end{tabular}
\end{table*}

\begin{table*}[h]
  \setlength{\abovecaptionskip}{0.1 cm}
    \centering
    \caption{The attack success rates (\%) of different methods based on SIM~\cite{lin2019nesterov} and TAIG-R~\cite{huang2022transferable}. ``-'' indicates the white-box attack.}
    \label{tab:opt}
    \begin{tabular}{@{~}c@{~~}c@{~~}c@{~~}c@{~~}c@{~~}c@{~~}c@{~~}c@{~~}c@{~~}c@{~~}c@{~~}c@{~}}
        \toprule
        Surrogate Model & Method & Average & Inc-v3 & Inc-v4 & IncRes-v2 & Inc-v3$_{ens3}$ & Inc-v3$_{ens4}$ & IncRes-v2$_{ens}$& HGD & R\&D & NIPS-r3\\
         \cmidrule(r){1-1}
         \cmidrule(r){2-2}
         \cmidrule(r){3-3}
         \cmidrule(r){4-12}
          \multirow{6}*{Inc-v3} & SIM & 37.7 & - & 72.1 & 71.4 & 37.3 & 36.1 & 20.2 & 16.0 & 19.7 & 28.9  \\
          & GN-SIM & 33.6 & - & 74.4 & 72.2 & 30.6 & 29.0 & 15.7 & 9.5 & 15.4 & 22.3 \\
          & MUP-SIM & \textbf{42.8} & - & \textbf{83.6} & \textbf{83.0} & \textbf{41.8} & \textbf{40.5} & \textbf{21.8} & \textbf{17.3} & \textbf{22.1} & \textbf{32.4} \ \\
          \cmidrule(r){2-12}
          & TAIG-R & 55.6 & - & 86.0 & 84.1 & 59.0 & 57.9 & 37.8 & 31.7 & 37.8 & 50.1 \\
         & GN-TAIG-R & 50.9 & - & 89.5 & 87.6 & 51.9 & 51.5 & 31.0 & 23.1 & 30.7 & 41.7  \\
         & MUP-TAIG-R & \textbf{60.5} & - & \textbf{92.1} & \textbf{90.4} & \textbf{63.7} & \textbf{63.9} & \textbf{40.7} & \textbf{34.1} & \textbf{42.8} & \textbf{56.7} \\
        \midrule
          \multirow{6}*{Inc-v4} & SIM & 45.6 & 83.9 & - & 75.4 & 47.4 & 42.5 & 27.2 & 22.6 & 28.2 & 37.8  \\
          & GN-SIM & 34.5 & 81.9 & - & 74.1 & 30.3 & 28.6 & 15.7 & 8.0 & 15.3 & 22.4 \\
          & MUP-SIM & \textbf{53.2} & \textbf{92.9} & - & \textbf{90.0} & \textbf{56.6} & \textbf{52.7} & \textbf{31.8} & \textbf{25.7} & \textbf{32.3} & \textbf{43.8} \\
          \cmidrule(r){2-12}
         & TAIG-R & 61.4 & 89.3 & - & 86.2 & 64.8 & 61.9 & 47.4 & 37.1 & 48.0 & 56.3   \\
          & GN-TAIG-R & 49.5 & 93.8 & - & 88.0 & 50.1 & 45.7 & 29.1 & 17.2 & 29.9 & 42.3  \\
          & MUP-TAIG-R & \textbf{66.6} & \textbf{94.9} & - & \textbf{94.0} & \textbf{72.1} & \textbf{68.7} & \textbf{50.2} & \textbf{38.4} & \textbf{51.6} & \textbf{62.9} \\
        \midrule
        \multirow{4}*{DN121} & SIM & 67.8 & 83.1 & 78.5 & 73.8 & 68.3 & 67.5 & 55.1 & 66.6 & 57.2 & 60.4  \\
        & MUP-SIM & \textbf{74.0} & \textbf{90.3} & \textbf{88.7} & \textbf{84.2} & \textbf{74.0} & \textbf{71.4} & \textbf{57.5} & \textbf{73.7} & \textbf{60.5} & \textbf{65.5}  \\
          \cmidrule(r){2-12}
         & TAIG-R & 85.7 & 91.9 & 92.1 & 88.8 & 87.1 & 85.6 & 77.2 & 86.7 & 79.2 & 82.5  \\
         & MUP-TAIG-R & \textbf{88.6} & \textbf{95.3} & \textbf{94.9} & \textbf{92.7} & \textbf{89.1} & \textbf{88.4} & \textbf{80.1} & \textbf{89.9} & \textbf{81.5} & \textbf{85.4}    \\
        \bottomrule
    \end{tabular}
\end{table*}

We list the experimental results in Table~\ref{tab:MIM}. The case where the surrogate model is the same as the victim model is not considered, as this is a white-box attack that we are not interested in. We also report the average attack success rates in Table~\ref{tab:MIM}, which is averaged over all nine victim models for DN121 and over eight victim models other than itself for Inc-v3 and Inc-v4. From the results, we can see that the MUP-MIM significantly outperforms MIM in terms of attack success rates in all experiments. Specifically, when Inc-v3, Inc-v4, and DN121 are used as surrogate models, MUP-MIM improves the attack success rate by 5.1\%, 5.8\%, and 11.3\%, respectively.

MUP and GN can also be compared based on the results in Table~\ref{tab:MIM}. From it we can see that MUP-MIM outperforms GN-MIM in all experiments when Inc-v3 is used as the surrogate model. However, GN-MIM performs slightly better than MUP-MIM in attacking IncRes-v2$_{ens}$, R\&D, and NIPS-r3 when Inc-v4 is used as the surrogate model. This indicates that the performance of MUP and GN are comparable when using MIM to generate adversarial examples and makes us curious about the comparison results between them when assisting an advanced optimizer.

Therefore, we further conduct experiments to compare the compatibility of GN and MUP with more optimizers. We considered SIM and TAIG-R and list the experimental results in Table~\ref{tab:opt}. Surprisingly, although GN performs well in assisting the MIM optimizer as claimed by~\cite{li2020learning}, the results in Table~\ref{tab:opt} show that using GN is worse than not using it when assisting more advanced optimizers. For example, when using Inc-v3 as a surrogate model, GN decreases the attack success rates of SIM and TAIG-R optimizer by 1.1\% and 4.7\%, respectively, on average. This drop is even more significant when using Inc-v4 as a surrogate model. On the contrary, our MUP assists the SIM and TAIG-R well, bringing further improvements in terms of attack success rates in all experiments. As a result, compared to TAIG-R, using MUP improves the attack success rates by 4.9\%, 4.2\%, and 2.9\% on average respectively when Inc-v3, Inc-v4, and DN121 are used as surrogate models. In addition, MUP improves the attack success rates of DN121 on nine victim models to at least \textbf{80.1\%} (IncRes-v2$_{ens}$). It should be pointed out that a high attack success rate is what is of interest in the field of transfer-based attack, since the goal of a malicious adversary is to fool a DNN system with as high a probability as possible. Thus, our MUP is more practical in real-world attack scenarios, although it is not always better than the GN when using a relatively weak optimizer (MIM).

\subsection{Ablation Studies}
In this subsection, we first conduct experiments to investigate the effect of the masking ratio $r$. Then, we have considered another metric for calculating the parameter importance scores, to highlight the effectiveness of the Taylor expansion-based metric now in use.

\begin{figure*}[h]
  \setlength{\abovecaptionskip}{0.1 cm}
  \centering
  \subfigure
    {\includegraphics[width=2.34in]{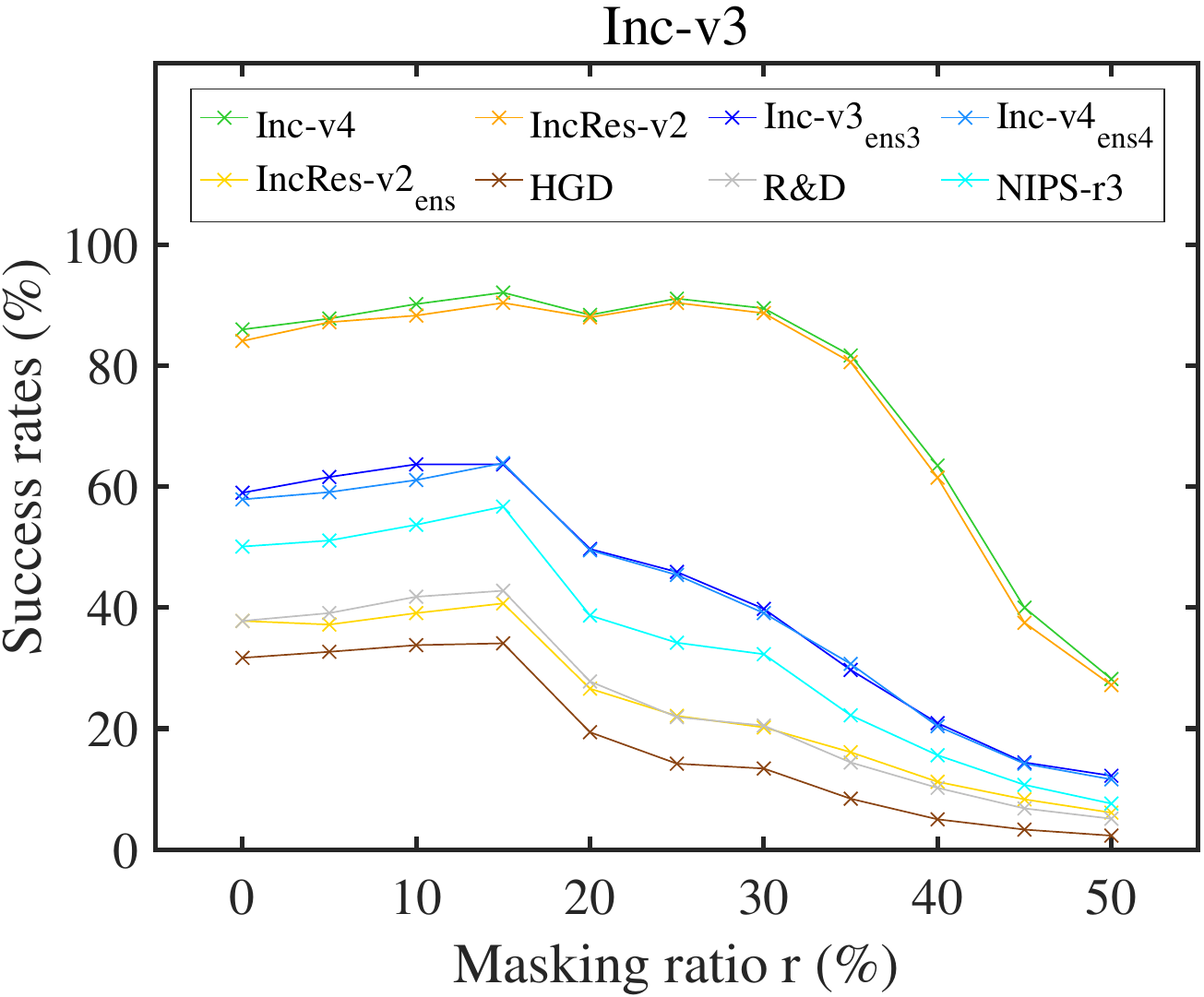}}
    \subfigure
    {\includegraphics[width=2.34in]{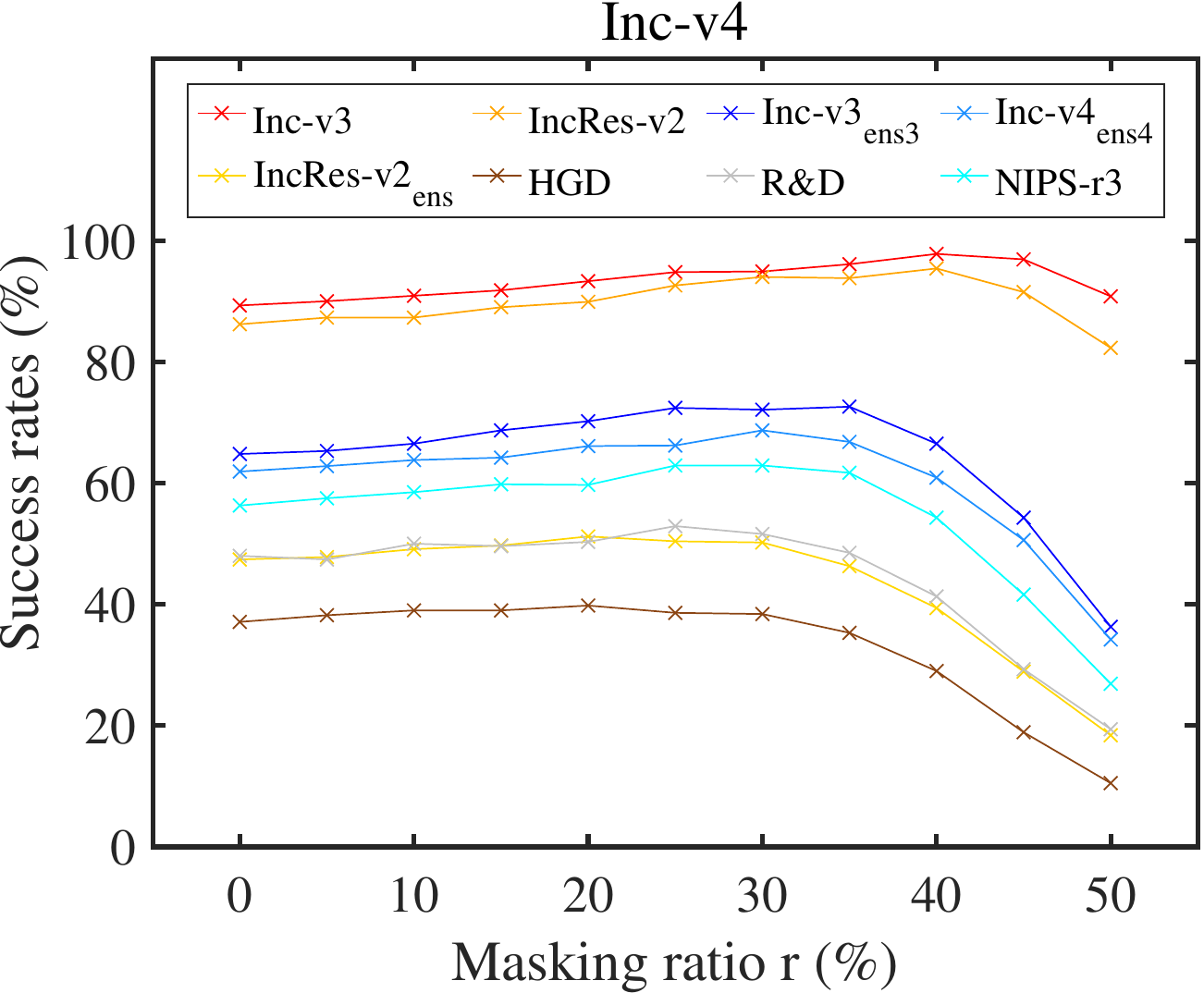}}
    \subfigure
    {\includegraphics[width=2.34in]{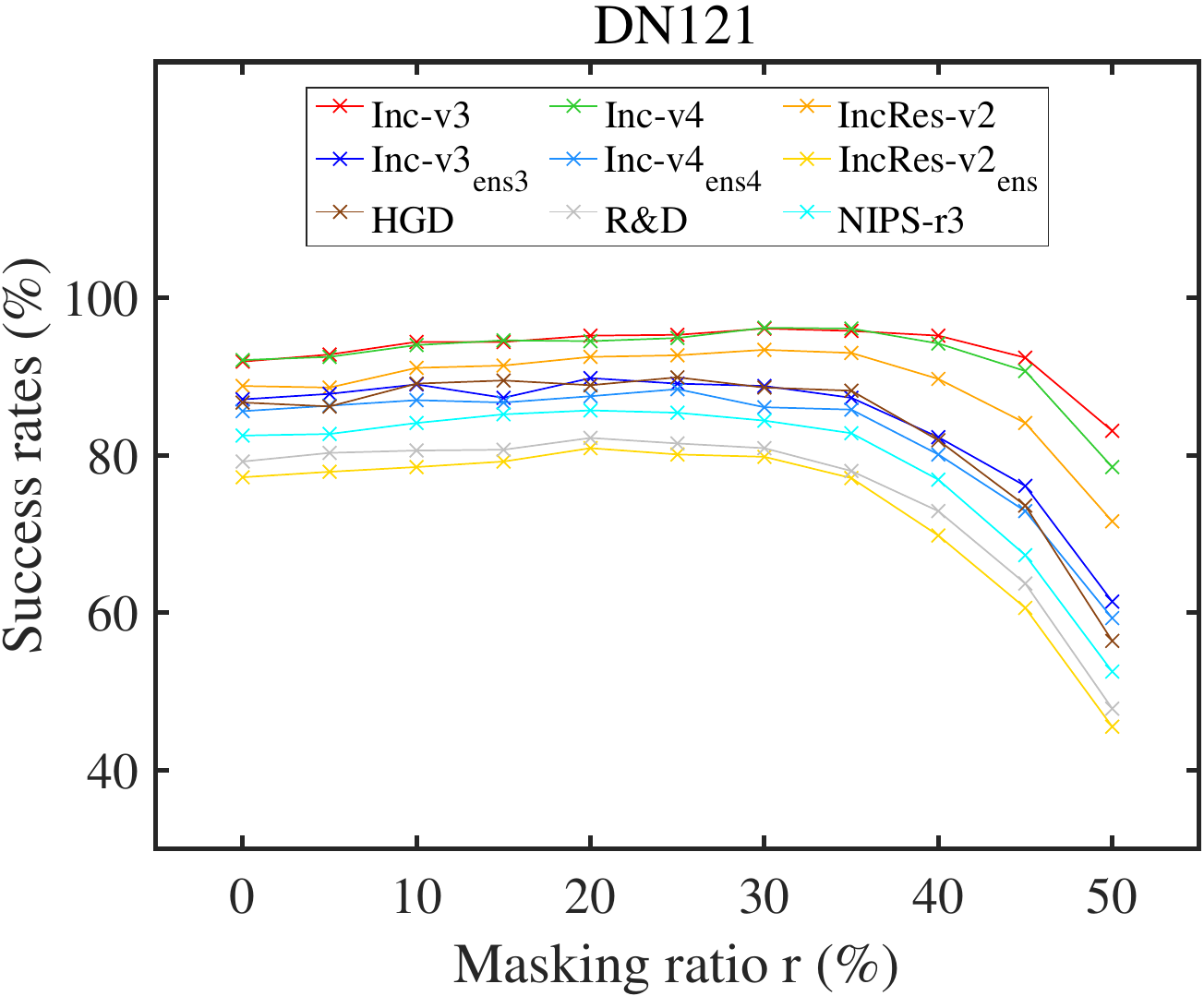}}
    \caption{The attack success rates (\%) versus hyperparameter $r$ curves. The surrogate models are Inc-v3, Inc-v4 and DN121 from left to right.}
      \label{fig:ratio}
\end{figure*}

The masking ratio $r$ plays an important role in MUP. When $r$ is equal to zero, it means that no parameters are masked. On the other hand, if $r$ is too large, most of the parameters of the surrogate model will be set to zero, which will impair the accuracy of the model prediction and lead to poor transferability. A natural question to ask is how should the hyperparameter $r$ be set. Will the optimal masking ratio $r$ vary greatly for different victim models? To answer this question, we enumerate different $r$ from $\{0\%, 5\%, 10\%, \cdots, 45\%, 50\%\}$ and conduct experiments on Inc-v3, Inc-v4 and DN121, respectively. We fix the optimizer for generating adversarial examples as TAIG-R since we are more interested in the performance of MUP to assist a stronger optimizer. The experimental results of using different masking ratios to assist TAIG-R are shown in Fig.~\ref{fig:ratio}, which shows that for the same surrogate model, the curves of attack success rates versus masking ratio $r$ are similar for all different victim models. Specifically, all experiments show that the attack success rate will initially increase, then reach the saturation region, and begin to decline rapidly. Moreover, \textit{the saturation region does not depend on the victim model}, because all curves of the same surrogate model have highly overlapping saturation regions. Therefore, determining the hyperparameter $r$ of MUP on a new surrogate model requires only choosing any public model as the victim model to search for an optimal $r$. This is critical for the transfer-based attack because the adversary has no knowledge of the victim model.

\begin{figure}[h]
  \setlength{\abovecaptionskip}{0.1 cm}
  \centering
    \includegraphics[width=2.1in]{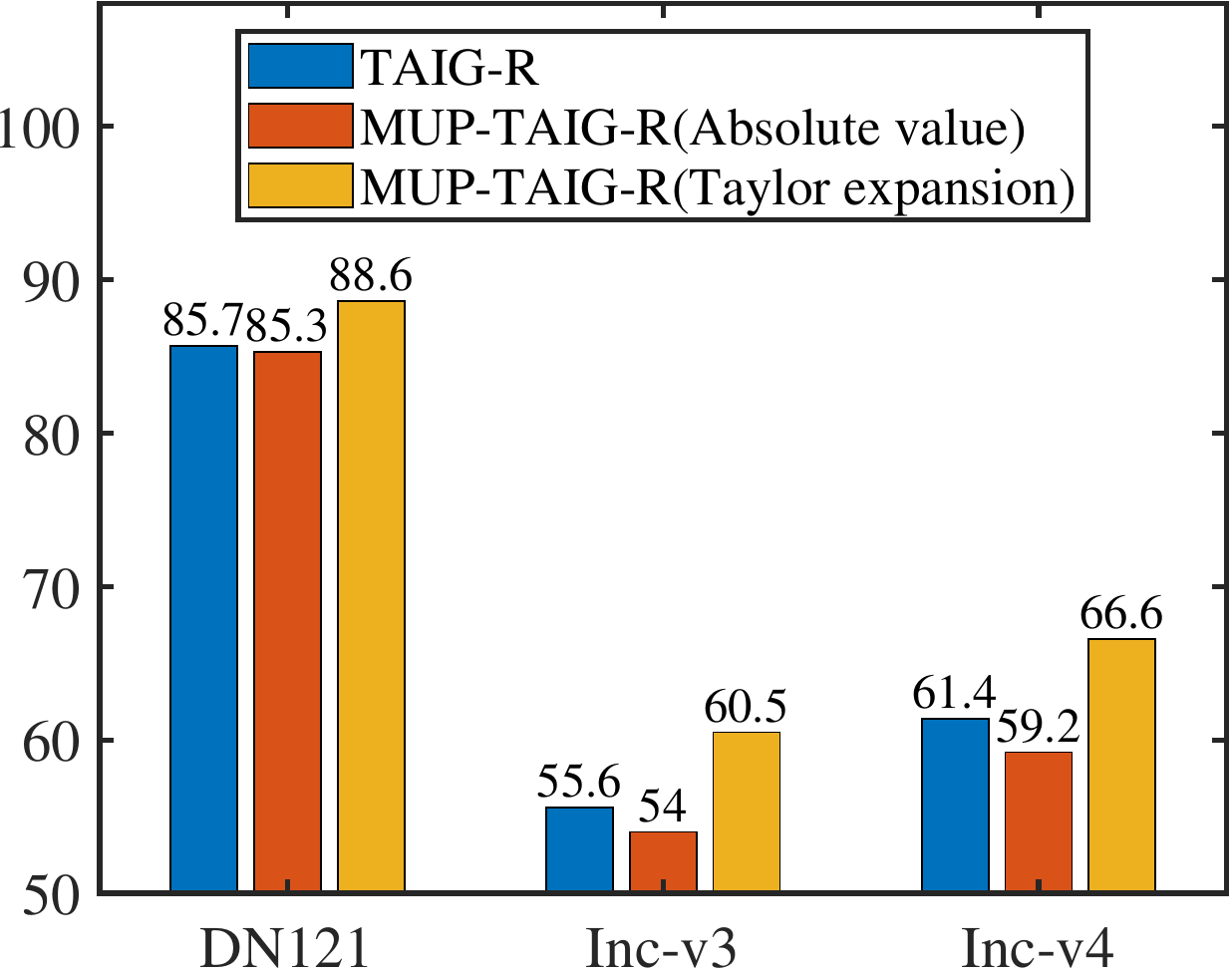}
  \caption{The average attack success rate (\%) for different methods using surrogate models Inc-v3, Inc-v4 and DN121, respectively.
  }
    \label{fig:bar}
\end{figure}

Finally, we additionally considered a metric for calculating the parameter importance scores, which is simply using the absolute value of the parameters as the importance scores, namely $V_i=|\theta_i|$. This metric was previously applied in model pruning by Han et al.~\cite{han2016deep}. By iteratively removing the parameters with the smallest absolute value, they compressed a VGG16~\cite{simonyan2014very} model by 13X with no loss of accuracy. By comparing it with the Taylor expansion-based metric (see~\eqref{eq:eq1}), the only difference is that the latter has a gradient term $\frac{\partial \mathcal{L}(x,y,\theta)}{\partial \theta_i}$. We use the absolute value-based metric to perform the MUP algorithm with the same hyperparameter $r$ as the MUP with the Taylor expansion-based metric, i.e., set to 15\%, 30\%, and 25\% Inc-v3, Inc-v4, and DN121, respectively. The experimental results of using different metrics to assist TAIG-R are shown in Fig.~\ref{fig:bar}, which shows that MUP with the absolute value-based metric does not boost the transfer-based attack. We believe this is because the absolute value-based metric is input-independent and ignores the features of inputs. In contrast, the Taylor expansion-based metric compensates for this shortcoming by multiplying an input-dependent gradient term. This indicates that the metric for calculating the importance scores is crucial for the proposed MUP-based methods.
\section{Conclusions}
In this paper, we propose the idea of improving the transferability of the adversarial examples by masking unimportant parameters (MUP). Based on this idea, we propose a technique for boosting transfer-based attacks, which utilizes a Taylor expansion-based metric to evaluate the parameter importance scores, and improves existing optimizers for the transfer-based attack by masking unimportant parameters. Extensive experiments verify the effectiveness of the proposed method.
\section{Acknowledgment}
This work was supported by the National Key Research and Development Plan of China (2020AAA0103502).
\bibliographystyle{IEEEtran}
\bibliography{IEEEexample}

\end{document}